\documentclass[]{beingbeyond}
\usepackage{enumitem}
\usepackage[toc,page,header]{appendix}

\usepackage[utf8]{inputenc} 
\usepackage[T1]{fontenc}    
\usepackage{hyperref}       
\usepackage{url}            
\usepackage{array}          
\usepackage{booktabs}       
\usepackage{amsfonts}       
\usepackage{nicefrac}       
\usepackage{microtype}      
\usepackage{xcolor}         
\usepackage{xspace}
\usepackage{bm}
\usepackage{bbm}
\usepackage{tabularx}
\usepackage{amssymb}
\usepackage{enumitem}
\usepackage{amsmath}
\usepackage{mathtools}
\usepackage{amsthm}
\usepackage{multirow}
\usepackage{makecell}
\usepackage{color}
\usepackage{colortbl}
\usepackage{adjustbox}
\usepackage{caption}
\usepackage{graphicx}
\usepackage{pifont}
\usepackage{wrapfig}
\usepackage{mdframed}
\usepackage{multicol}

\newcommand{\method}{SENTINEL\xspace}

\title{SENTINEL: A Fully End-to-End Language-Action Model for Humanoid Whole Body Control}

\author{{\bfseries Yuxuan Wang$^{1,2*}$ \quad Haobin Jiang$^{1,2*}$ \quad Shiqing Yao$^{2}$ \quad Ziluo Ding$^{2}$ \quad Zongqing Lu$^{1,2,\dagger}$}}

\affiliation{{$^{1}$Peking University \quad $^{2}$BeingBeyond}}

\abstract{
Existing humanoid control systems often rely on teleoperation or modular generation pipelines that separate language understanding from physical execution. However, the former is entirely human-driven, and the latter lacks tight alignment between language commands and physical behaviors. In this paper, we present \textbf{\texttt{\method}}, a fully end-to-end language–action model for humanoid whole-body control. We construct a large-scale dataset by tracking human motions in simulation using a pretrained whole body controller, combined with their text annotations. The model directly maps language commands and proprioceptive inputs to low-level actions without any intermediate representation. The model generates action chunks using flow matching, which can be subsequently refined by a residual action head for real-world deployment. Our method exhibits strong semantic understanding and stable execution on humanoid robots in both simulation and real-world deployment, and also supports multi-modal extensions by converting inputs into texts.
}

\checkdata[Video Page]{\url{https://youtu.be/U5B5Pgw1N3A}}
\checkdata[Date]{November 24, 2025}

\begin{document}

\maketitle

\begingroup
\renewcommand\thefootnote{\fnsymbol{footnote}} 
\setcounter{footnote}{0}
\footnotetext[1]{These authors contributed equally to this work.}
\footnotetext[2]{Correspondence to Zongqing Lu $<$lu@beingbeyond.com$>$.}
\endgroup

\section{Introduction}
Humanoid robots are envisioned to serve as versatile embodied agents capable of performing diverse tasks in human environments. Yet, most existing humanoids are still controlled through joysticks \citep{li2025amo,cheng2024expressive,ben2025homie} or tele-operation interfaces \citep{ji2024exbody2,fu2024humanplus,he2025omnih2o}, which requires expert operation and lack flexibility. To make humanoids truly accessible and intelligent, language control is a natural direction, allowing users to specify actions and intentions in an intuitive, high-level manner. Language offers a rich and scalable interface that can describe complex motion goals and temporal structures, making it a core pathway toward expressive and generalizable humanoid behavior.

A straightforward approach to realizing language-conditioned humanoid motion is to couple a text-to-motion generation model \citep{tevet2023human, zhang2024motiondiffuse,zhang2023generating, jiang2024motiongpt} with a downstream whole body controller (WBC) \citep{wang2025experts,chen2025gmt,liao2025beyondmimic}. The text-to-motion module generates motion sequences conditioned on language descriptions, while the controller executes them on the humanoid. However, such a decoupled design often fails to produce physically feasible motions, as the two components are optimized independently. Currently, there are two possible solutions to address this issue: (i) introducing an alignment module that aligns the output of the motion generation model with the input space of the WBC, ensuring that the generated representations can be effectively interpreted and executed by the controller \citep{xue2025leverb,mao2025universal,shao2025langwbc}, or (ii) developing an \textbf{\texttt{end-to-end model}} that directly maps language to physically feasible humanoid motor control without any intermediate representation. This paradigm offers stronger semantic–kinematic consistency and smoother execution. Unlike modular pipelines with hand-crafted intermediate motion representations or physical constraints, the end-to-end approach enables gradient flow from execution feedback back to language understanding, allowing joint optimization across the entire system. While it is more challenging to train, successfully realizing this approach can ultimately lead to more robust, generalizable, and scalable language-conditioned embodied intelligence.

Building on this motivation, we introduce \textbf{\texttt{\method}}, a fully end-to-end framework that translates natural language instructions into low-level whole body control for humanoids. To train the model, we first construct a language-action dataset by using a whole body controller to track human motion paired with natural language descriptions in simulation. This is followed by a pre-training stage, where an end-to-end language–action model with a flow-matching \citep{lipman2023flow,liu2023flow} action expert is trained to predict humanoid actions from the humanoid’s proprioceptive state and the language command.
After pre-training, we perform a post-training refinement stage using residual reinforcement learning \citep{johannink2019residual,yuan2025policy} in simulation with domain randomization. This stage further mitigates the open-loop drift introduced by action chunking and adapts the model for real-world deployment. In addition, \textbf{\texttt{\method}} naturally supports multi-modal extensions by converting other sensory inputs into language control signals; for example, visual observations can be translated into waypoint targets and integrated into the model, enabling the humanoid to navigate by combining language instructions with perceptual cues.

To assess its language grounding and control capabilities, we evaluate \textbf{\texttt{\method}} in simulation and on real humanoid hardware. Our experiments demonstrate that the model handles a wide range of language-conditioned tasks, including locomotion and complex whole body movements. In real-world deployment, \textbf{\texttt{\method}} exhibits reliable zero-shot sim-to-real transfer, producing physically stable and feasible actions. Quantitative and qualitative results confirm that our approach achieves stronger semantic grounding and more robust motion execution than existing baselines. To the best of our knowledge, \textbf{\texttt{\method}} is the first fully end-to-end language–action model for humanoid control that operates without intermediate motion representations or latent abstractions.

\section{Related Work}
\noindent \textbf{Human Motion Generation.} 
High-quality open-source human motion datasets~\citep{AMASS:ICCV:2019, li2023object, Guo_2022_CVPR, zhang2025motion, lin2023motionx} provide a solid foundation for human motion generation. With the rapid development of deep learning, motion generation have progressed substantially in both fidelity and diversity. Within this line of work, text-to-motion generation has emerged as a major research direction. Early approaches formulated the task in a sequence-to-sequence manner~\citep{Guo_2022_CVPR, petrovich23tmr, petrovich2022temos}, while subsequent diffusion-based methods~\citep{tevet2023human, zhang2024motiondiffuse, chen2023executing, kim2023flame, yuan2023physdiff} significantly increase the quality and diversity of generated motions. Another paradigm is to use discrete token to represent human motion and generate motion sequences in an autoregressive progress~\citep{guo2022tm2t, zhang2023generating, jiang2024motiongpt, li2023enhancing}.

\noindent \textbf{Humanoid Whole Body Control.} DeepMimic~\citep{peng2018deepmimic, bergamin2019drecon, Park:2019} introduced a reinforcement learning framework that trains a policy to control humanoid avatars to mimic a single reference human motion. Subsequent works~\citep{peng2021amp, Luo2023PerpetualHC, tessler2024maskedmimic} extended the framework to multi-task settings while relaxing the requirement for motion quality. However, these methods remain limited to virtual characters and overlook the structural gaps between human and humanoid robots. Meanwhile, humanoid robots have undergone rapid development in recent years. \citep{he2025omnih2o, fu2024humanplus} were among the first to propose using retargeting methods to transfer human motions to humanoid robots and successfully reproduced the same motion patterns on real robots. \citep{ze2025twist, he2024learning, fu2025demohlm, li2025amo} enabled teleoperation by tracking human motion using a motion capture system. However, these methods rely on high-quality human motion input and robust low-level control policies.

\noindent \textbf{Text-Based Whole Body Control.} Recent works have started to control the humanoid robot by language input. UH-1~\citep{mao2025universal} uses an open-loop motion generation approach, ignoring environmental feedback and risking failure in real world. LangWBC~\citep{shao2025langwbc} distills expert policies via a CVAE \citep{sohn2015learning}, but its MLP-based design limits both expressiveness and robustness. \citep{ding2025humanoid, xue2025leverb, kalaria2025dreamcontrol} use a hierarchical modular architecture to fuse vision, language, and motion. However, all these works, although not directly using motion, still require intermediate representations to realize the final frameworks. In contrast, our method leverages a transformer-based architecture integrated with a flow-matching action expert, enabling the system to both interpret semantic instructions and respond promptly to physical feedback for maintaining balance in an end-to-end manner.

\section{Preliminary}
\label{section:RL}

\begin{figure}[t]
    \centering
    \includegraphics[width=0.9\linewidth]{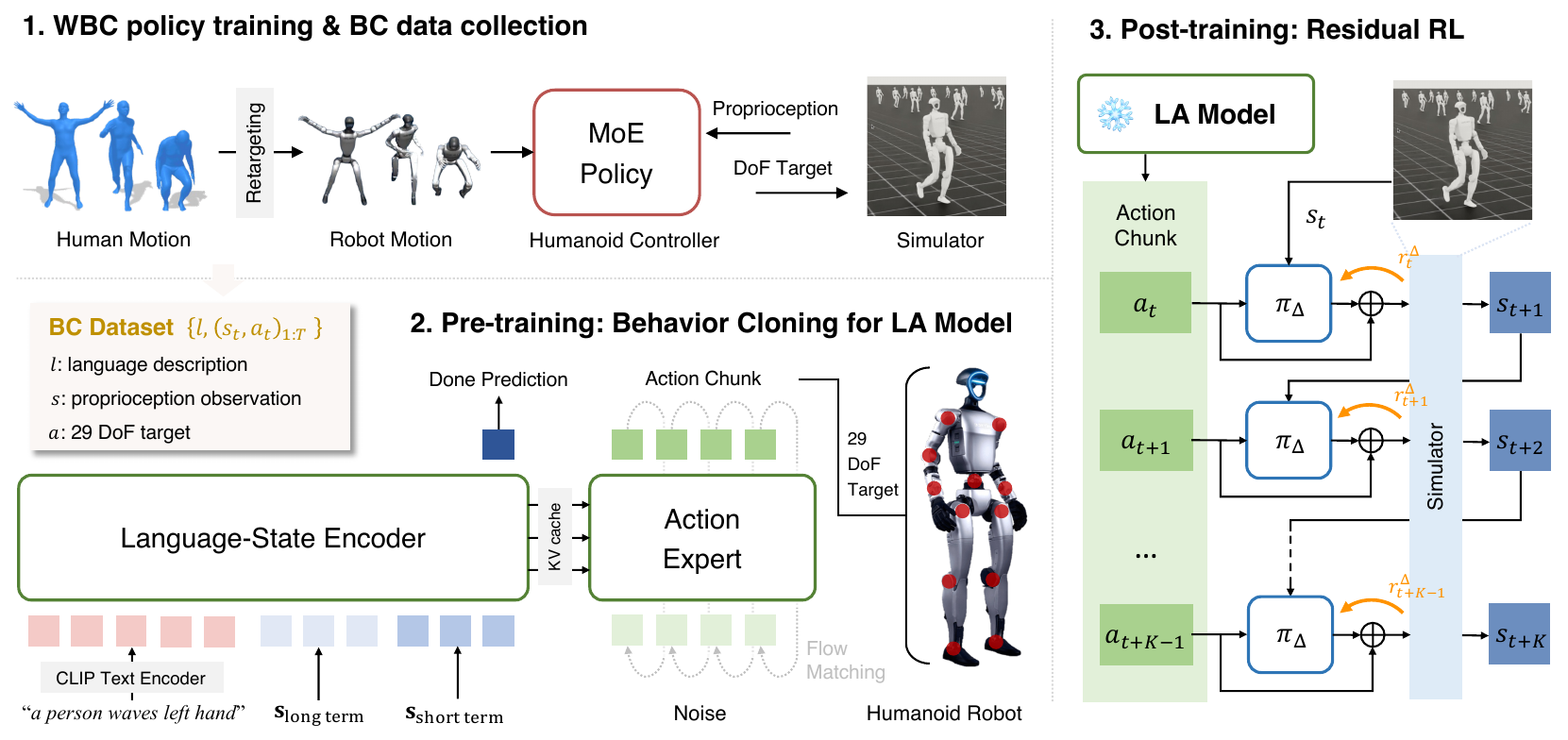}
    \caption{\textbf{Overview of \textbf{\texttt{\method}}}. Our framework consists of three stages. (1) We construct a language-action dataset by using a whole body controller to track human motion data paired with natural language descriptions. (2) We train an end-to-end language–action model with flow matching action head, which predicts a robot action chunk conditioned on both the proprioceptive state history and the language command. (3) A post-training stage with a residual action head is introduced to enhance its performance.}
    \label{fig:method}
\end{figure}

For humanoid robots such as the Unitree G1, whole body control aims to compute target joint commands that enable the robot to follow a desired motion while consistent with its dynamics and physical constraints. Whole body control is commonly formulated as a goal-conditioned reinforcement learning problem~\citep{he2025omnih2o,ji2024exbody2,ding2025jaeger,wang2025experts,chen2025gmt}, where an policy is trained to output target joint positions. In this setting, the policy receives two types of inputs: a motion-tracking target $m_t^{\rm ref}$ and the robot’s proprioceptive state $s_t$.

The target $m_t^{\rm ref}$ is obtained by retargeting human motion (\textit{e.g.}, SMPL format~\citep{SMPL:2015}) into the robot’s kinematic space. Human motions are acquired from publicly available motion datasets \citep{AMASS:ICCV:2019, li2023object, Guo_2022_CVPR, lin2023motionx} or generated via text-to-motion models \citep{tevet2023human, zhang2024motiondiffuse,zhang2023generating, jiang2024motiongpt}. The proprioceptive state $s_t$ includes the robot's root linear velocity $v^{\rm root}_t \in \mathbb{R}^3$, root angular velocity $\omega^{\rm root}_t \in \mathbb{R}^3$, projected gravity vector $g_t \in \mathbb{R}^3$, joint positions $q_t \in \mathbb{R}^{29}$, joint velocities $\dot{q}_t \in \mathbb{R}^{29}$, and the previous action $a_{t-1} \in \mathbb{R}^{29}$:
\begin{equation}
    s_t = [v^{\mathrm{root}}_t, \omega^{\mathrm{root}}_t, g_t, q_t, \dot{q}_t, a_{t-1}].
    \label{equ:state}
\end{equation}
The policy output \(a_t\) is linearly projected to compute the target joint positions as
$
q^{\text{target}}_t = a_t \cdot \alpha + \bar{q},
$
where \(\alpha\) denotes the scale parameters and \(\bar{q}\) represents a constant nominal joint configuration, following the convention in~\citep{liao2025beyondmimic}. Note that the policy is not intended to directly copy the retargeted motion $m_t^{\rm ref}$, but rather to learn a control strategy that adapts this reference motion to the robot's current physical state and the laws of physics. The representation of $m_t^{\rm ref}$ and other training details are provided in \Cref{app:wbc}.

\section{Method}

In \Cref{fig:method}, we illustrate the overall framework of our method, \textbf{\texttt{\method}}. We begin by constructing a language–action dataset, built upon existing human motion datasets~\citep{AMASS:ICCV:2019, Guo_2022_CVPR} and a whole body control policy trained via reinforcement learning \citep{wang2025experts,liao2025beyondmimic} (\Cref{subsec:data}). We then present our end-to-end language–action model based on flow matching \citep{lipman2023flow,liu2023flow}, and subsequently describe a post-training stage that employs a residual action head trained with reinforcement learning (\Cref{subsec:train}). Finally, our framework supports multi-modal extensions by converting additional sensory inputs into language commands (\Cref{subsec:vision}).

\subsection{Data Curation}
\label{subsec:data}

To bridge the gap between human motion and humanoid control, we first train a whole body controller capable of tracking diverse human motions on humanoid robots, as introduced in \Cref{section:RL}. We adopt a Mixture-of-Expert (MoE) policy and train it via PPO \citep{schulman2017proximal} in IsaacLab \citep{mittal2023orbit}. We then leverage this controller to generate a large-scale language-action dataset by collecting robot execution trajectories that correspond to human motion data paired with natural language descriptions.

\label{sec:robot_data}

Given a language-labeled human motion dataset $\mathcal{D}_{\mathcal{M}} = \{(\mathbf{m}_i, l_i)\}_{i=1}^N$, where $\mathbf{m}_i$ represents a human motion sequence and $l_i$ is the associated natural language description. For each human motion sequence $\mathbf{m}_i$, we rollout the whole body controller in the simulator to produce a robot trajectory $\boldsymbol{\tau}_i$ that tracking the retargeted human motion. This results in a new dataset $\mathcal{D}_{\rm robot} = \{(\boldsymbol{\tau}_i, l_i)\}_{i=1}^N$ which pairs physically-grounded robot trajectories with natural language descriptions. Specifically, the trajectory $\boldsymbol{\tau}_i$ consists of a sequence of states and actions: $\boldsymbol{\tau}_i = \{(s_t, a_t)\}_{t=1}^{T_i}$, where $s_t$ denotes the robot's state at time step $t$ and $a_t$ denotes the action taken, as defined in \Cref{section:RL}.

During data collection, we enable domain randomization to broaden the data coverage. The randomization introduces variations in physical properties, \textit{e.g.}, center of mass and friction coefficients, as well as external perturbations, including random pushes and noise in torque computation. These randomized factors perturb the collected trajectories, thereby enhancing the robustness and generalization of the model trained on this dataset \citep{truong2024pdp}.

\subsection{Model Training}
\label{subsec:train}

In this section, we detail the architecture and training procedure of our end-to-end language–action model. The model predicts robot actions conditioned on both the proprioceptive state history and a natural language command, utilizing flow matching \citep{lipman2023flow,liu2023flow} for effective modeling. Our model directly predicts low-level actions for humanoid whole body control without a intermediate motion representation, enabling seamless integration of high-level language inputs and low-level control execution.

\subsubsection{Model Architecture}

Following recent advances in diffusion and flow matching action heads for vision-language-action (VLA) models \citep{black2024pi_0,intelligence2025pi_,shukor2025smolvla,bjorck2025gr00t}, our model contains two main components: a language–state encoder for contextual representation learning, and an action expert for robot action prediction.

As illustrated in \Cref{fig:method}, the language–state encoder processes the contextual input $c_t$, consisting of the natural language command $l$ and the robot’s state history $\mathbf{s}_t^{\rm hist}$:
\begin{equation}
    c_t = \left[l, \mathbf{s}_t^{\rm hist} \right].
\end{equation}
The language tokens are first obtained from a CLIP text encoder~\citep{radford2021learning}, producing semantic token-level embeddings of the command. In parallel, the model receives the robot’s state history, where each state includes proprioceptive observations as defined in \Cref{equ:state}. The language–state encoder is a Transformer \citep{vaswani2017attention} that jointly processes these inputs to learn a unified contextual representation, capturing temporal dependencies within the state sequence and cross-modal interactions between language and state. The representations are exposed to the action expert via a key and value cache interface for subsequent action generation.

The action expert is implemented as a flow matching model~\citep{lipman2023flow,liu2023flow}, which predicts an action chunk $\mathbf{A}_t = [a_t, a_{t+1}, \cdots, a_{t+H-1}]$ for the next $H$ steps conditioned on the encoded context. Formally, the action expert $v_\theta$ is trained to approximate the ground-truth velocity field of the action flow conditioned on the context $c_t$. Given a ground-truth action chunk $\mathbf{A}_t$ and a random interpolation coefficient $\tau \sim \mathrm{Beta}(1.5, 1.0)$, we construct a noisy sample $\mathbf{A}_t^\tau = \tau \epsilon + (1 - \tau) \mathbf{A}_t$ with $\epsilon \sim \mathcal{N}(0, I)$, which represents an intermediate point along the transport path between the Gaussian prior and the target action manifold. The training objective minimizes the distance between the predicted velocity field and the target velocity field:
\begin{equation}
\label{equ:fm}
    L(\theta) = \mathbb{E}_{p(\mathbf{A}_t | c_t), q(\mathbf{A}_t^\tau | \mathbf{A}_t)} \left\| v_\theta(\mathbf{A}_t^\tau, \tau, c_t) - u(\mathbf{A}_t^\tau | \mathbf{A}_t) \right\|^2,
\end{equation}
where the target velocity is defined as $u(\mathbf{A}_t^\tau | \mathbf{A}_t) = \epsilon - \mathbf{A}_t$. 

The action expert is implemented as a Transformer decoder that attends to the cached keys and values from the language–state encoder, enabling the generation of actions that are contextually aligned with both the robot’s state history and the natural language command.
Following \citep{shukor2025smolvla}, we adopt the same architectural backbone as the language–state encoder but with a reduced hidden dimension, and interleave cross-attention and self-attention layers to provide high-quality action predictions while maintaining efficient inference speed.

During inference, we integrate the predicted velocity field over discrete time steps $\Delta t$ (0.1 in our experiments) from $\tau=1$ to $\tau=0$, following the learned flow:
\begin{equation}
\mathbf{A}_t^{\tau - \Delta t}
= \mathbf{A}_t^{\tau} - v_\theta(\mathbf{A}_t^{\tau}, \tau, c_t) \cdot \Delta t.
\end{equation}
This integration process provides the action chunk $\mathbf{A}_t$ for the next $H$ steps. At execution time, only the first $K$ actions are applied to the robot, after which a new chunk is generated in a receding-horizon manner.

Model details and hyperparameter settings for training and inference are provided in \Cref{app:model}.

\vspace{1mm}
\noindent \textbf{Multi-Scale Observation.} Unlike most manipulation tasks, where task information and progress can be inferred from the current visual observation, humanoid whole body control requires reasoning over longer temporal horizons to capture motion dynamics and intent. For example, to execute a command such as “\textit{walk forward four steps and stop.}”, the model must recognize how many steps have already been taken and plan the remaining steps accordingly. Relying solely on recent states may lead to shortsighted actions that fail to satisfy the command. 

To address this, we construct a multi-scale state history that combines fine-grained short-term states and coarse-grained long-term states: 
\begin{equation}
    \label{equ:state_input}
    \mathbf{s}_t^{\rm hist} = [\mathbf{s}_t^{\rm long\_term}, \mathbf{s}_t^{\rm short\_term}].
\end{equation}
Specifically, $\mathbf{s}_t^{\rm short\_term}$ contains the most recent 10 time-steps sampled at 50 Hz, providing high-resolution proprioceptive feedback for stable control, and $\mathbf{s}_t^{\rm long\_term}$ contains states sampled at a lower frequency (4Hz) over the past 10 seconds, offering a broader temporal context to capture motion patterns and task progress. This multi-scale representation enables the model to balance short-term precision and long-term awareness during action generation.

\vspace{1mm}
\noindent \textbf{Dynamics-Aware Prediction.} 

In humanoid whole-body control, where balance and momentum are tightly encoded in the robot’s physical state, jointly predicting future states along with actions provides additional supervision that regularizes the policy toward physically grounded behavior. Relying on action prediction alone does not explicitly enforce consistency with the underlying dynamics.
To this end, we extend the action expert to predict an augmented action chunk $\Tilde{\mathbf{A}}_t = [\Tilde{a}_t,\Tilde{a}_{t+1},\cdots,\Tilde{a}_{t+H-1}]$, where
\begin{equation}
    \label{equ:augmented_action}
    \Tilde{a}_t = [a_t, v_{t+1}^{\rm root}, \omega_{t+1}^{\rm root}, q_{t+1}].
\end{equation}
This encourages the model to learn not only the mapping from context to actions but also the transition dynamics of the environment. Such auxiliary state-prediction objectives have been commonly adopted in reinforcement learning and behavior cloning \citep{janner2021offline,janner2022planning,huang2025diffuse}. Note that our state prediction differs from motion generation models, as it serves only as an auxiliary signal and does not train the model to reconstruct fine-grained joint trajectories or rotations.

\vspace{1mm}
\noindent \textbf{Done Prediction.} In text-based whole body control, it is crucial for the policy to determine when to terminate the current action sequence, since there is no fixed horizon or predefined stopping criterion as in manipulation tasks. Inspired by autoregressive models, we introduce a binary \texttt{done} token indicating whether the command is expected to be completed within the next $H$ steps. This token is predicted by an MLP head applied to the last hidden state of the final state token in the language–state encoder. During inference, if the predicted \texttt{done} probability exceeds the threshold (0.5) for consecutive $\lfloor H / K \rfloor$ chunks, the model actively terminates the current command execution.

\subsubsection{Residual Post-Training}
\label{subsec:posttrain}

To further mitigate open-loop drift caused by action chunking and improve sim-to-real transferability, we introduce a post-training stage that learns a lightweight residual action head $\pi_\Delta$.
Given the current state $s_t$ and the augmented action $\tilde{a}_t$ predicted by the action expert, $\pi_\Delta$ outputs a residual action:
\begin{equation}
    \Delta a_t = \pi_\Delta(s_t, \Tilde{a}_t).
\end{equation}
The final action applied to the robot is then computed as $a_t^{\rm final} = a_t + \Delta a_t$, allowing the model to adapt to unmodeled effects during behavior cloning, while preserving the semantic intention. In addition, the lightweight residual head provides timely feedback and adjustments for high-dynamics motions before the next action chunk is predicted.

\begin{table}[t]
\caption{Main reward terms for residual post-training.}
\label{tab:delta_reward}
\centering
\resizebox{0.58\linewidth}{!}
{
\begingroup

\setlength{\tabcolsep}{10pt} 
\renewcommand{\arraystretch}{1.3} 
\begin{tabular}{llc}
        \toprule
        \multicolumn{1}{c}{\textbf{Term}} & \multicolumn{1}{c}{\textbf{Expression}} & \textbf{Weight}  \\
        \midrule
        DoF tracking                & $\exp{(-\|q_t - \hat{q}_t\| / 0.09)}$       & 2.0 \\
        Residual action norm        & $\exp{(-\|\Delta a_t\| / 0.09)}$            & 2.0 \\
        Action rate                 & $-\|a_t^{\rm final} - a_{t-1}^{\rm final}\|$  & 0.2 \\
        Termination                 & $-1$ if fall down                             & 100. \\
        \bottomrule
    \end{tabular}
\endgroup
}
\end{table}

To train $\pi_\Delta$, we freeze the parameters of the language–action model and optimize the residual head with PPO \citep{schulman2017proximal} in IsaacLab \citep{mittal2023orbit}. We apply domain randomization over physical parameters and external perturbations to improve the robustness of our method. The reward functions consists of tracking terms and regularization terms, as summarized in \Cref{tab:delta_reward}. The tracking term encourages the residual action to drive the future joint positions toward the original predictions $\hat{q}$ under domain randomization. The regularization terms constrain the magnitude of the residual action and the action rate to ensure smooth and physically plausible actions. The full training details and reward terms are provided in \Cref{app:posttrain}.

\subsection{Modality Expansion}
\label{sec:Modality}

Our model supports multi-modal extensions by converting additional sensory inputs into language signals. For instance, navigation typically relies on visual perception, but can be formulated using spatial coordinates derived from off-the-shelf detection models \citep{liu2024grounding,wen2024foundationpose} for humanoid navigation and manipulation \citep{fu2025demohlm,kalaria2025dreamcontrol}. In this paper, we formulate navigation goals as language commands that specifies target waypoints in the robot’s egocentric frame, such as “\textit{a person walks to $(x\ m, y\ m)$}”. To enable this, we relabel the locomotion trajectories in the collected dataset $\mathcal{D}_{\mathrm{robot}}$ by discretizing the relative displacement between the start and end positions, thereby constructing a navigation-oriented subset $\mathcal{D}_{\mathrm{robot}}^{\mathrm{nav}}$. This dataset provides additional supervision for learning precise navigation and locomotion control.

\begin{figure}[t]
    \centering
    \includegraphics[width=0.65\linewidth]{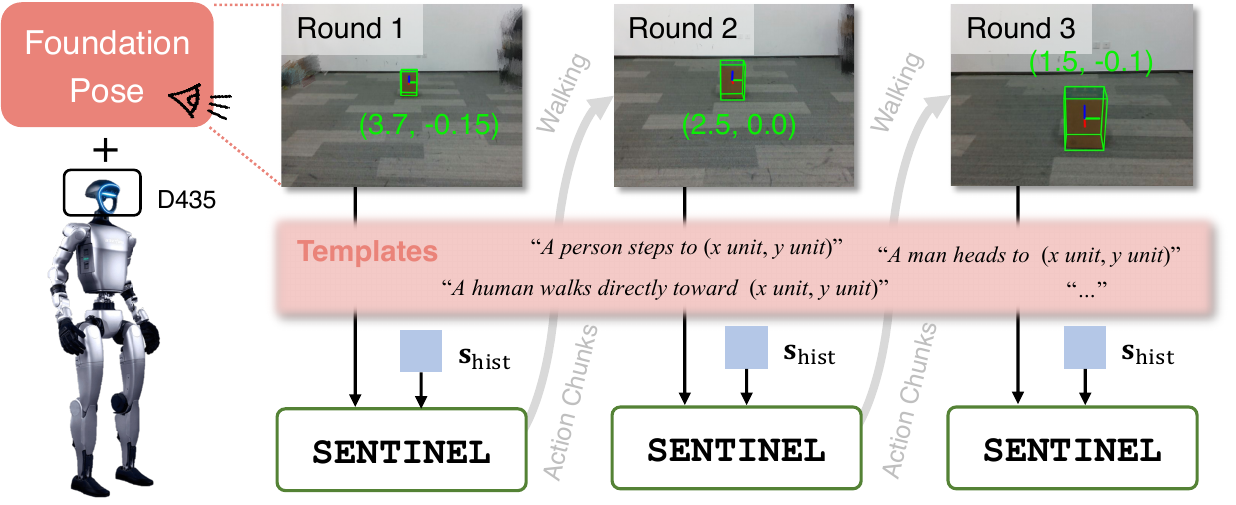}
    \caption{Integration of visual perception into \textbf{\texttt{\method}} for navigation tasks. The onboard D435 camera captures front-view RGB-D images, which are processed by FoundationPose~\citep{wen2024foundationpose} to estimate the target position in the robot’s egocentric frame. The estimated waypoint is then inserted into natural-language command templates and provided to \textbf{\texttt{\method}}, together with the robot’s proprioceptive state, to generate whole body control actions. This closed-loop process enables the robot to iteratively approach the visual target.}
    \label{fig:vision}
\end{figure}

As shown in \Cref{fig:vision}, in our real-world deployment, visual perception is incorporated through front-view RGB-D images captured by the onboard D435 camera. The images are processed using FoundationPose \citep{wen2024foundationpose} to estimate the relative position of a target object in the robot’s current coordinate frame. The estimated waypoint is then inserted into predefined natural-language command templates, which is fed into \textbf{\texttt{\method}} together with the robot’s proprioceptive state history for action generation. This process can be executed iteratively in a closed-loop manner, allowing the robot to refine its movement and gradually move toward the visual target.

\label{subsec:vision}

\section{Experiments}

\subsection{Experiment Setup}

\noindent \textbf{Dataset.} We conduct experiments on a text-labeled and PHC-filtered \citep{Luo2023PerpetualHC} subset of the AMASS dataset \citep{AMASS:ICCV:2019}.
This subset and its mirrored counterpart contain 12,422 motion sequences, each paired with three to four textual descriptions. On this dataset, we first train a whole body controller for the \textbf{Unitree G1} and then use the trained controller to collect a robot trajectory dataset $\mathcal{D}_{\mathrm{robot}}$ under multiple domain-randomization patterns, as described in \Cref{sec:robot_data}. In total, we collect approximately 200,000 language-labeled robot trajectories, corresponding to about 100 million state–action pairs. Additional details regarding the dataset composition are provided in \Cref{app:data}.

\vspace{1mm}
\noindent \textbf{Baselines.} We compare our method with the following baselines. (i) \textbf{MDM + Retarget} generates human motions from text using MDM \citep{tevet2023human}, and our trained whole body controller tracks the retargeted motion sequences on the robot. (ii) \textbf{T2M-GPT + Retarget} follows a similar two-stage pipeline, where T2M-GPT \citep{zhang2023generating} is used for text-conditioned human motion generation. These two methods represent the paradigm that combines a human motion generation model with a whole body controller for text-based whole body control. (iii) \textbf{UH-1} \citep{mao2025universal} learns to generate robot poses directly from text, which are subsequently tracked by the whole body controller without motion retargeting\footnote{The UH-1's text-to-action mode is not evaluated, as its open-loop control formulation does not yield executable trajectories in the simulation.}. (iv) \textbf{LangWBC} \citep{shao2025langwbc} trains a text-based whole body controller using a CVAE \citep{sohn2015learning} with DAgger~\citep{ross2011reduction} to imitate a teacher control policy. These two methods does not use human motion generation, but relies on other intermediate representations. More details about baseline implementations are provided in \Cref{app:baseline}.

\vspace{1mm}
\noindent \textbf{Evaluation Metrics.} We evaluate model performance using both generation quality metrics and physical execution success on the test set, following the HumanML3D division \citep{Guo_2022_CVPR}. Provided text prompts, we rollout the learned model in IsaacLab \citep{mittal2023orbit} and collect robot trajectories. For generation evaluation, we train a text–motion retrieval (TMR) model~\citep{petrovich23tmr} on $\mathcal{D}_{\rm robot}$ as the feature extractor for better language-motion alignment. 
We reports Multi-Modal Distance (\textbf{MM-Dist}), R-precision at K (\textbf{R@K}, K=1,2,3), \textbf{Diversity}, and Maximum Mean Discrepancy (\textbf{MMD}) \citep{gretton2012kernel}, as detailed in \Cref{app:metrics}. We do not report Frechet Inception Distance (FID) \citep{heusel2017gans,Guo_2022_CVPR}, since it assumes the embedding features follow a Gaussian distribution, which does not hold for contrastively trained encoders such as TMR. Instead, we adopt MMD with a Gaussian RBF kernel to provide a more reliable distance measure \citep{jayasumana2024rethinking}. For physical plausibility, we report the success rate of executing generated actions in the simulator, where success is defined as the robot maintaining balance without falling before timeout or active termination via the done prediction mechanism.

\subsection{Text-Based Whole Body Control Evaluation}

\begin{table}[t]
\caption{Text-based whole body control evaluation results. Our method outperforms all baselines in both generation quality and physical execution success. $\rightarrow$ means the closer to Ground Truth the better.}
\label{tab:main_result}
\centering
\resizebox{1.0\textwidth}{!}
{
\begingroup

\setlength{\tabcolsep}{8pt} 
\renewcommand{\arraystretch}{1.3} 
\begin{tabular}{lccccccc}
\toprule
\textbf{Method} & MM-Dist $\downarrow$ & R@1 $\uparrow$ & R@2 $\uparrow$ & R@3 $\uparrow$ & Diversity $\rightarrow$ & MMD ($1e^{-2}$) $\downarrow$ & Success Rate (\%) $\uparrow$ \\
\midrule
\textbf{Ground Truth} & 0.110 & 0.969 & 0.995 & 0.999 & 0.977 & - & - \\
\midrule
\textbf{MDM + Retarget} & 0.703 & 0.338 & 0.470 & 0.559 & 0.928 & 8.910 & 94.94 \\
\textbf{T2M-GPT + Retarget} & 0.577 & 0.481 & 0.637 & 0.714 & 0.957 & 4.115 & 89.33 \\
\textbf{UH-1} & 0.644 & 0.394 & 0.525 & 0.585 & 0.950 & 4.729 & 86.95 \\
\textbf{LangWBC} & 0.682 & 0.435 & 0.565 & 0.622 & 0.828 & 8.642 & 81.78 \\
\textbf{\textbf{\method}} (ours)&
\textbf{0.487} & 
\textbf{0.582} & 
\textbf{0.717} & 
\textbf{0.766} & 
\textbf{0.967} &
\textbf{3.438} & 
\textbf{99.45} \\
\bottomrule
\end{tabular}
\endgroup
}
\end{table}

The main results are summarized in \Cref{tab:main_result}. First, we aim to answer the question: \textbf{\textit{Does training on physically grounded robot data enhance text-based whole body control?}} Methods based on human motion generation, such as MDM and T2M-GPT, have significantly advanced text-to-motion generation in the human motion domain \citep{tevet2023human,zhang2023generating,jiang2024motiongpt,zhang2024motiondiffuse}. However, when the generated motions are retargeted and executed through the whole body controller, their lack of physical constraints and the absence of exposure to real robot dynamics result in suboptimal control performance. UH-1, while bypassing motion retargeting by generating robot poses directly, still relies on retargeted pose data rather than trajectories collected through real or simulated interaction. In contrast, our method outperforms these baselines by incorporating physically grounded robot trajectories into the training process, leading to improved generation quality and execution success. Such end-to-end training on state and action data collected from real dynamics extend the text-to-motion generation paradigm toward physically grounded robot control.

\begin{wrapfigure}{r}{0.50\textwidth}
    \centering
    \vspace{-5mm} 
    \includegraphics[width=\linewidth]{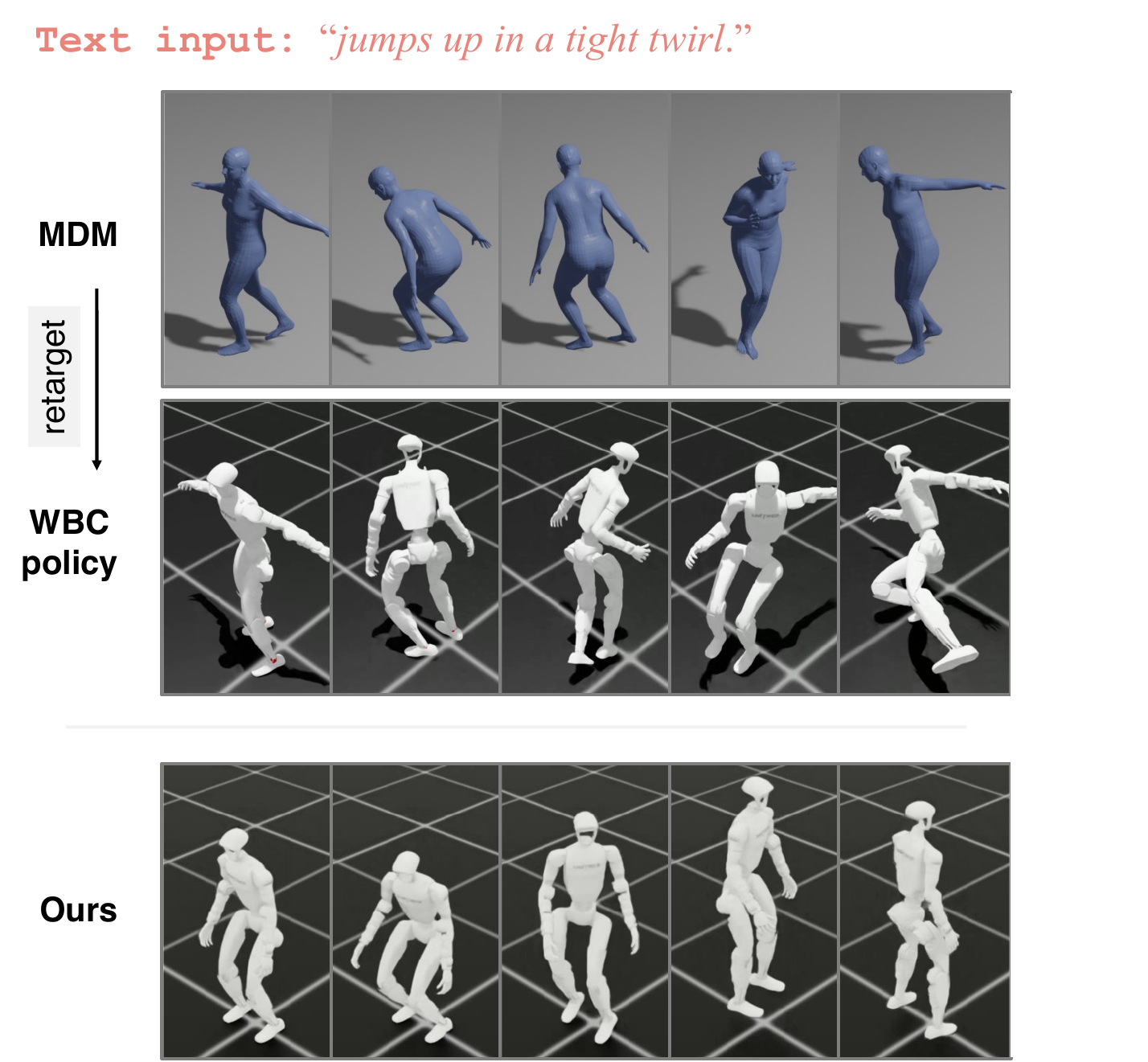}
    \caption{Comparison between MDM + Retarget \citep{tevet2023human} and our method on an example text prompt: “\textit{jumps up in a tight twirl.}”.}
    \label{fig:compare}
\end{wrapfigure}

Second, we investigate: \textbf{\textit{Is a large transformer necessary for effective text-based whole body control?}} Different from other baselines, LangWBC employs a MLP-based architecture to learn the text-to-action mapping. While it could achieve physical fidelity through DAgger in simulation, its limited model capacity results in lower language alignment and poor control generalization to unseen commands in the test set. In contrast, our transformer-based model effectively captures both fine-grained semantic understanding and physical dynamics, leading to superior performance across all evaluation metrics.

\vspace{1mm}
\noindent \textbf{Qualitative Comparison.} We present a qualitative comparison between our method and MDM + Retarget \citep{tevet2023human} in \Cref{fig:compare} for illustration. Given the text prompt “\textit{jumps up in a tight twirl.}”, MDM generates a human motion that captures the meaning but exhibits a large-amplitude spinning motion with rapid angular changes. When this motion is retargeted and executed by the whole body controller, the large-amplitude rotation cause the controller to lose balance, resulting in unstable execution and falling. In contrast, our method directly predicts physically feasible action chunks at the control level, producing a more moderated jumping–twirling behavior.

\subsection{Ablation Studies}

\begin{figure}[t]
    \centering
    \begin{minipage}{0.48\columnwidth}
        \captionof{table}{Ablation study results for model design.}
        \label{tab:ablation}
        \centering
        
        \resizebox{1.0\linewidth}{!}{
        \begingroup
        \setlength{\tabcolsep}{7pt}
        \renewcommand{\arraystretch}{1.4}
        \begin{tabular}{lccc}
        \toprule
        \textbf{Method} & R@1 $\uparrow$ & MMD $\downarrow$ & Success $\uparrow$ \\
        \midrule
        \textbf{Base} & 0.582 & \textbf{3.438} & \textbf{99.45} \\
        \textbf{w/ 0.2s Observation} & 0.153 & 72.468& \textbf{99.50}\\
        \textbf{w/ 2.0s Observation} & 0.489 & 3.956 & \textbf{99.56} \\
        \textbf{w/o State Prediction} & \textbf{0.589} & 3.587 & 98.67 \\
        \textbf{w/o Done Prediction} & 0.522 & 5.852 & \textbf{99.44} \\
        \bottomrule
        \end{tabular}
        \endgroup
        }
    \end{minipage}
    \quad
    \begin{minipage}{0.48\columnwidth}
        \begin{subfigure}{0.48\textwidth}
            \centering
            \includegraphics[width=\linewidth]{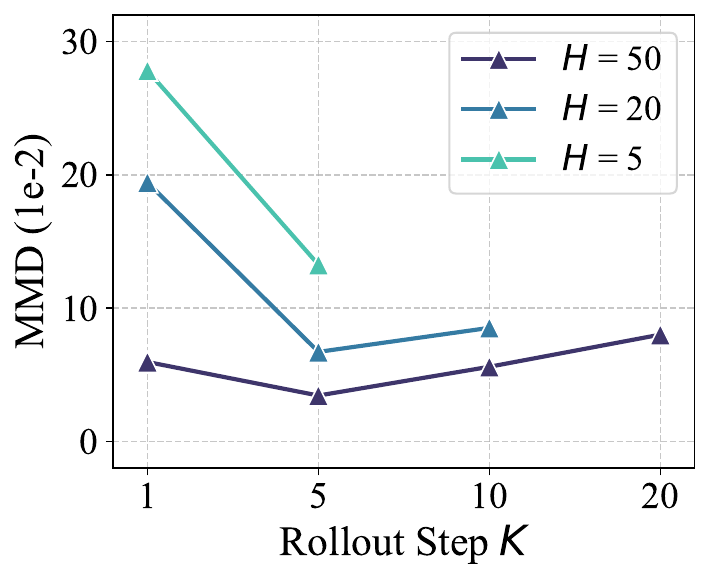}
            \vspace{-4mm}
            \caption{MMD $\downarrow$}
        \end{subfigure}
        \hfill
        \begin{subfigure}{0.49\textwidth}
            \centering
            \includegraphics[width=\linewidth]{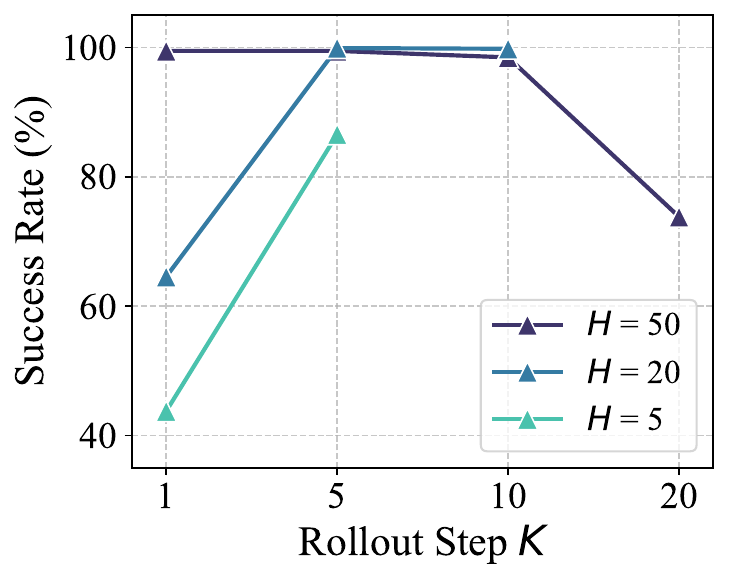}
            \vspace{-4mm}
            \caption{Success Rate $\uparrow$}
        \end{subfigure}
        \vspace{-1mm}
        \caption{Results for different action chunk sizes $H$ (training) and rollout steps $K$ (inference).}
        \label{fig:chunk_length}
    \end{minipage}
\end{figure}

\noindent \textbf{Model Design.} 
In this ablation study, we investigate the contributions of different components in our model architecture, as described in \Cref{subsec:train}. Specifically, we evaluate the performance impact of the following modifications: (i) \textbf{w/ 0.2s Observation}: removing the long-term observation history $\mathbf{s}_t^{\rm long\_term}$ from the model input (\Cref{equ:state_input}); (ii) \textbf{w/ 2.0s Observation}: replacing the observation with a 20-frame history at 10 Hz, consistent with LangWBC \citep{shao2025langwbc}; (iii) \textbf{w/o State Prediction}: removing the state prediction from the action expert outputs (\Cref{equ:augmented_action}); and (iv) \textbf{w/o Done Prediction}: removing the done prediction from the model. We conduct separate experiments for each modification while keeping all other settings unchanged.

The ablation results are presented in \Cref{tab:ablation}. When the long-term observation is removed, the model loses awareness of its temporal progress within a motion sequence, often resulting in premature termination prediction and expressive degradation. Without state prediction, the model achieves comparable performance but exhibits slightly lower MMD and success rate, indicating that state prediction enhances the execution stability by providing auxiliary supervision on future physical states.
Finally, removing the done prediction prevents the model from active termination during inference, leading to undesirable behaviors after completing the language command.

\vspace{1mm}
\noindent \textbf{Action Chunk Size.} The action chunk size $H$ determines how many future actions and states the model predicts during training, while the rollout step $K$ controls how many actions in the action chunk are executed before re-querying the model during inference. We conduct experiments with different combinations of $H$ and $K$ to assess their impact on performance, and the results are summarized in \Cref{fig:chunk_length}. Across all action chunk sizes, a rollout step of $K = 5$ consistently provides the best performance in both semantic alignment and physical execution. Importantly, this does not imply that choosing $H = K$ is sufficient. Our results indicate that larger action chunk sizes yield significant benefits: performance improves steadily from $H = 5$ to $50$. Even though only a portion of the predicted chunk is executed at each inference step, training with a longer horizon encourages the model to learn longer-range semantic dependencies and maintain greater physical consistency. Therefore, we use $H = 50$ and $K = 5$ as the default configuration for our method.

\vspace{1mm}
\noindent \textbf{Model Scaling.} To assess the impact of model size, we train and evaluate three variants of our model with different parameter counts: a large model with 600 million parameters (our default setting), a medium model with 200 million parameters, and a small model with 60 million parameters. All models are trained under the same settings and evaluated using the same metrics. The results are summarized in \Cref{tab:model_size}. The results indicate that larger models consistently outperform smaller ones across all evaluation metrics. From 60M to 600M parameters, there is a significant improvement in R@1 (from 0.099 to 0.582), MMD (from 0.139 to 0.034), and success rate (from 22.73\% to 99.45\%). This trend highlights the importance of model capacity in capturing the complex relationships between language commands, robot states, and physical dynamics necessary for effective text-based whole body control.

\subsection{Post-Training Evaluation}

In this section, by evaluating in the simulator with domain randomization, we aim to answer: \textbf{\textit{Is the post-training with a residual action head effective for text-based whole body control?}} We conduct experiments on the base model and perform post-training using the residual action head. We apply domain randomization and set rollout step $K=10$ to better evaluate the effectiveness of the residual network, which introduces physical and time latency redundancy for real-world deployment. The results are summarized in \Cref{tab:posttrain}, indicating that the base model struggles with both fidelity and stability in more realistic, high-variability environments. Meanwhile, significant improvements in R@1 (from 0.315 to 0.392) and success rate (from 95.44\% to 99.11\%) are observed with the assistance of the residual action head. This shows that the residual action head effectively mitigates the drift caused by action chunking and enhances the model’s sim-to-real transferability, thereby enabling more robust text-based whole-body control.

\begin{figure}[t]
    \centering
    \begin{minipage}{0.45\columnwidth}
        \captionof{table}{Results for different model sizes.}
        \label{tab:model_size}
        \centering
        
        \resizebox{\linewidth}{!}{
        \begingroup
        \setlength{\tabcolsep}{10pt}
        \renewcommand{\arraystretch}{1.3}
        \begin{tabular}{lccc}
        \toprule
        \textbf{Model Size} & R@1 $\uparrow$ & MMD $\downarrow$ & Success $\uparrow$ \\
        \midrule
        \textbf{600M} & \textbf{0.582} & \textbf{3.438} & \textbf{99.45} \\
        \textbf{200M} & 0.371 & 6.485 & 99.09 \\
        \textbf{60M}  & 0.099 & 13.968 & 22.73 \\
        \bottomrule
        \end{tabular}
        \endgroup
        }
    \end{minipage}
    \quad
    \begin{minipage}{0.45\columnwidth}
        \centering
        \captionof{table}{Results for residual post-training.}
        \label{tab:posttrain}
        
        \resizebox{\linewidth}{!}{
            \begingroup
            \setlength{\tabcolsep}{8pt} 
            \renewcommand{\arraystretch}{1.3}
            \begin{tabular}{clcc}
                \toprule
                \textbf{Domain Rand} & \textbf{Method} & R@1 $\uparrow$ & Success $\uparrow$ \\
                \midrule
                \ding{55} & \textbf{Base} & 0.490 & 98.45 \\
                \midrule
                \ding{51} & \textbf{Base} & 0.315 & 95.44 \\
                \ding{51} & \textbf{Base + $\pi_{\Delta}$} & \textbf{0.392} & \textbf{99.11} \\
                \bottomrule
            \end{tabular}
            \endgroup
        }
    \end{minipage}
\end{figure}

\begin{figure}[t]
    \centering
    \vspace{2mm}
    \includegraphics[width=0.55\linewidth]{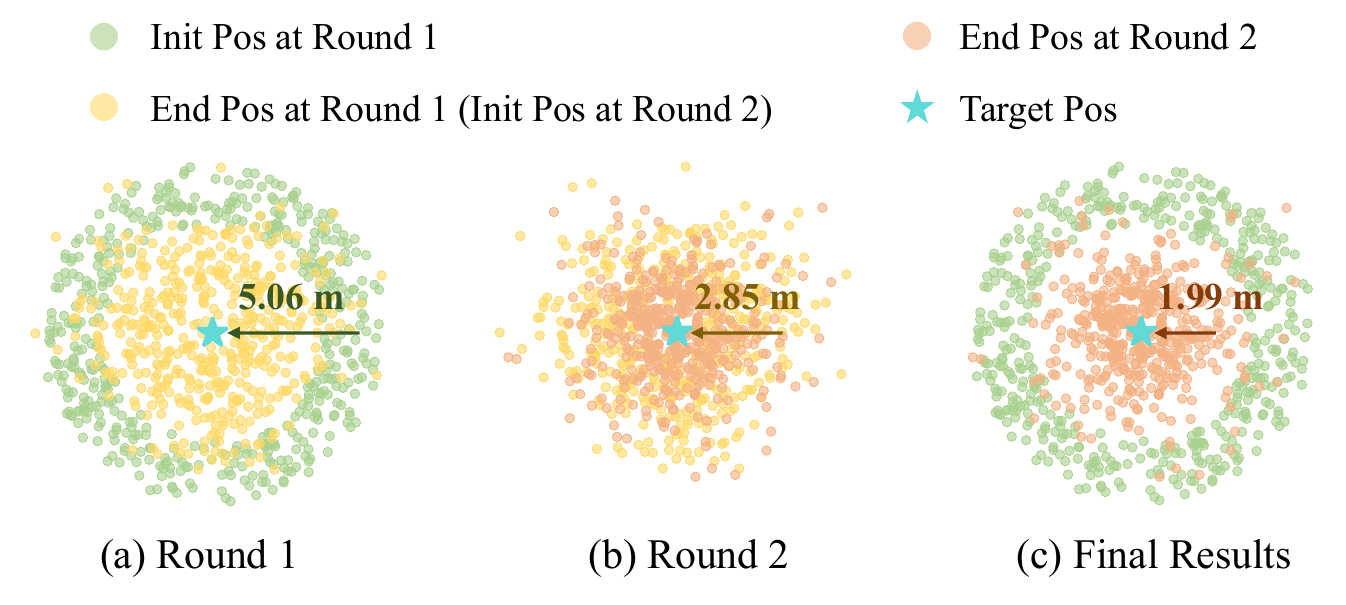}
    \caption{Visualization of robot positions during navigation iterations. To clearly illustrate the navigation progress, all target waypoints are translated to the center. The average distance decreases from 5.06 m to 2.85 m (Round 1) and further to 1.99 m (Round 2).}
    \label{fig:wp-result}
\end{figure}

\subsection{Waypoint Navigation Evaluation}
To evaluate whether \textbf{\texttt{\method}} can follow the navigation commands, as depicted in \Cref{sec:Modality}, we randomly generate 500 target waypoints with $x$-coordinates ranging from 4 m to 6 m and $y$-coordinates from -2 m to 2 m, which are then combined with pre-defined templates. The policy performs two rounds of navigation, as detailed in \Cref{app:navi}. As shown in \Cref{fig:wp-result}, the robot's positions gradually approach the target positions (5.06m $\rightarrow$ 2.85m $\rightarrow$ 1.99m). This further demonstrates that \textbf{\texttt{\method}} is able to perform the navigation tasks given positional inputs and can be easily extended to other modalities.

\subsection{Real-World Deployment} 

\begin{figure}[t]
    \centering
    \includegraphics[width=0.6\linewidth]{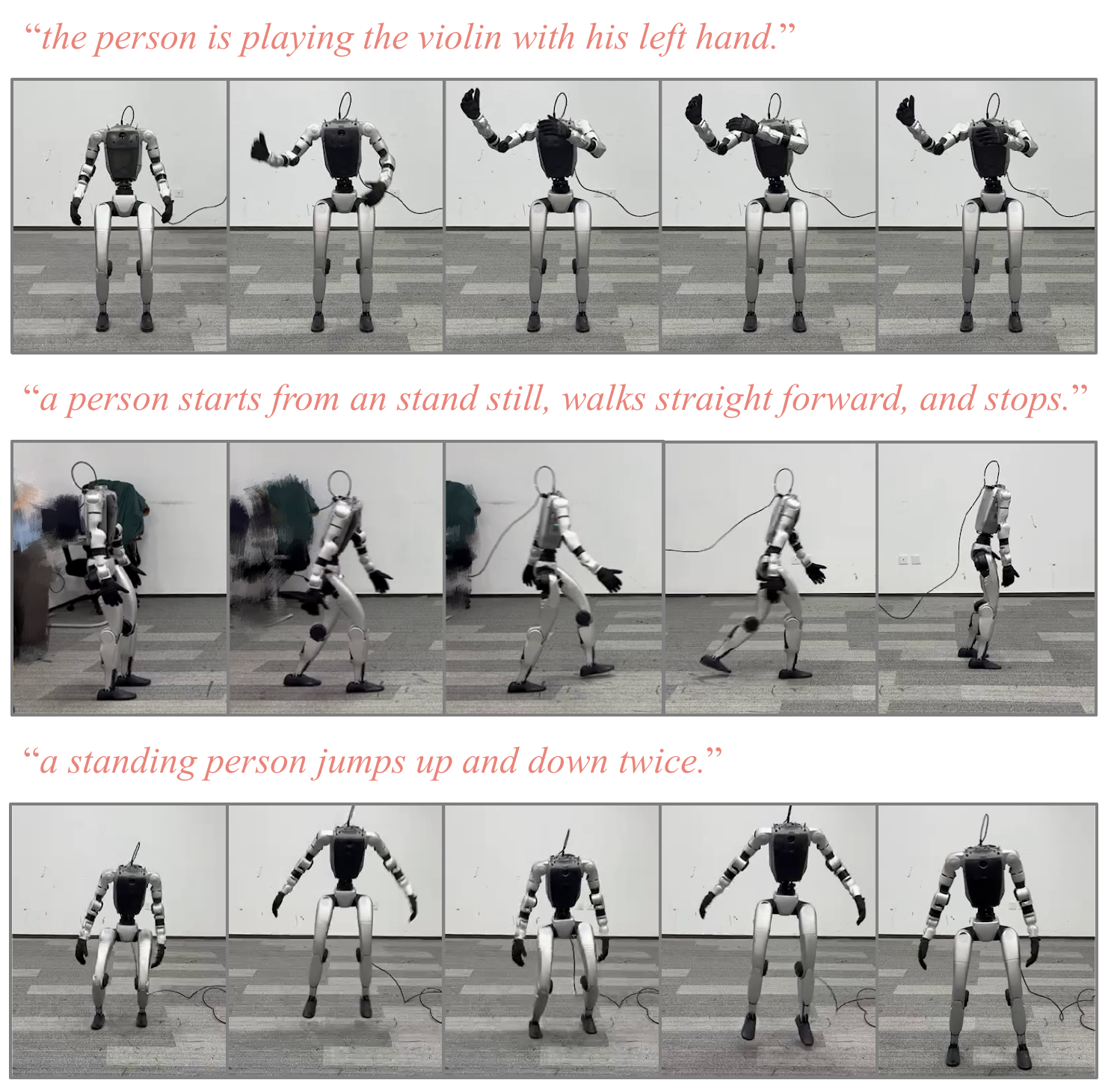}
    \caption{Real world deployment of our method on a Unitree G1 for text-based whole body control.}
    \label{fig:demo}
\end{figure}

We deploy our model, after residual post-training, on a Unitree G1 robot for real-world text-based whole body control, as shown in \Cref{fig:demo}. The robot is able to correctly interpret and execute a diverse set of commands, including upper-body behaviors (e.g., \textit{playing the violin}), locomotion (e.g., \textit{walking straight forward}), and more complex whole body motions (e.g., \textit{jumping up and down}). These successful executions in the real world demonstrate the effectiveness of our approach in bridging language understanding with physically grounded whole body control. Further details on real-world deployment and the asynchronous action-chunk inference pipeline are provided in \Cref{app:real}.

\section{Conclusion}

In this paper, we present \textbf{\texttt{\method}}, a fully end-to-end language–action model for text-based humanoid whole body control. By directly mapping natural language and proprioceptive inputs to low-level actions, the model generates semantically aligned and physically coherent behaviors without relying on intermediate representations. We further evaluate \textbf{\texttt{\method}} on real humanoid hardware and conduct waypoint-navigation experiments to assess its multimodal extensibility. These results highlight \textbf{\texttt{\method}} as a promising direction for scalable, general-purpose humanoid control driven by natural language.

\clearpage

\bibliographystyle{unsrt}
\bibliography{ref}

\clearpage

\beginappendix
\crefalias{section}{appendix}
\crefalias{subsection}{appendix}

\section{Language-Action Dataset}
\label{app:dataset}

\subsection{Humanoid Hardware}

This work uses the \textbf{Unitree G1}, a lightweight humanoid robot equipped with 29 actuated degrees of freedom (DoF): 7 DoF for each arm, 6 DoF for each leg, and 3 Dof at the waist. The robot uses a standard low-level joint-space Proportional–Derivative (PD) controller, which computes motor torques $\tau$ according to:
\begin{equation}
    \tau = K_p (q^{\text{target}} - q) - K_d \dot{q},
\end{equation}
where \(q\in\mathbb{R}^{29}\) and \(\dot{q}\in\mathbb{R}^{29}\) denote the measured joint angles and velocities, and \(q^{\text{target}}\in\mathbb{R}^{29}\) is the desired target joint position commanded by a policy controller.  The PD controller runs at 200 Hz, and the control policy runs at 50 Hz.

\subsection{Whole Body Control Tracking Policy}
\label{app:wbc}

We define the reference motion $m_t^{\rm ref}$ for tracking as:
\begin{equation}
\label{equ:target}
    m_t^{\rm ref} = \left[ x_t^{\mathrm{ref}}, r_t^{\mathrm{ref}}, q_t^{\mathrm{ref}}, \dot{q}_t^{\mathrm{ref}}, p_t^{\mathrm{ref}} \right],
\end{equation}
where \(x_t^{\mathrm{ref}} \in \mathbb{R}^3\) denotes the root translation, 
\(r_t^{\mathrm{ref}} \in \mathrm{SO}(3)\) denotes the root rotation, 
\(q_t^{\mathrm{ref}}, \dot{q}_t^{\mathrm{ref}} \in \mathbb{R}^{29}\) denotes the joint positions and velocities, 
and \(p_t^{\mathrm{ref}} \in \mathbb{R}^{4\times3}\) denotes the keypoint positions of the wrists and ankles in the local body frame.

The tracking policy $\pi_{\rm track}$ takes as input the proprioceptive state $s_t$ and motion-tracking target $m_t^{\rm ref}$, which are define in \Cref{equ:state} and \Cref{equ:target}, respectively. The policy outputs \(a_t\), which is linearly projected to compute the target joint positions $q^{\text{target}}_t$ for PD controller:
\begin{equation}
    q^{\text{target}}_t = a_t \cdot \alpha + \bar{q},
\end{equation}
where \(\alpha\) denotes the scale parameters and \(\bar{q}\) represents a constant nominal joint configuration, following the convention in~\citep{liao2025beyondmimic}.

To train the whole body control tracking policy, we use PPO \citep{schulman2017proximal} as our RL algorithm and IsaacLab \citep{mittal2023orbit} as the training environment. We implement a Mixture-of-Expert (MoE) policy to enhance its capability. The training hyperparameters, domain randomization, and reward terms are listed in \Cref{tab:ppo}, \Cref{tab:dmr}, and \Cref{tab:rew}, respectively. Following \citep{wang2025experts}, we apply PHC \citep{Luo2023PerpetualHC} to filter the AMASS dataset \citep{AMASS:ICCV:2019}, resulting in 8,179 high-quality motion sequences for whole body control training.

\begin{figure}[t]
    \centering
    \begin{minipage}{0.4\columnwidth}
        \captionof{table}{Hyperparameters for tracking policy training.}
        \label{tab:ppo}
        \centering
        \resizebox{0.95\linewidth}{!}
        {
        \begingroup
        
        \setlength{\tabcolsep}{10pt} 
        \renewcommand{\arraystretch}{1.3} 
        \begin{tabular}{lc}
                \toprule
                \multicolumn{1}{c}{\textbf{Term}} & \multicolumn{1}{c}{\textbf{Value}} \\
                \midrule
                Num parallel envs       & 4096  \\
                Num steps per env       & 24    \\
                Learning epochs         & 5     \\
                Num mini\_batches       & 4     \\
                Discount $\gamma$       & 0.99  \\
                GAE $\lambda$           & 0.95  \\
                PPO clip ratio          & 0.2   \\
                Value loss coefficient  & 1.0   \\
                Entropy coefficient     & 0.005 \\
                Initial learning rate   & 1e-3\\
                Adaptive LR range       & 1e-5, 1e-2\\
                Desired KL              & 0.01 \\
                Max grad norm           & 1.0 \\
                Optimizer               & Adam \\
                MLP layers              & [512, 256, 128] \\
                Num experts             & 12 \\
                \bottomrule
            \end{tabular}
        \endgroup
        }
    \end{minipage}
    \quad
    \begin{minipage}{0.55\columnwidth}
        \centering
        \captionof{table}{Domain randomization for tracking policy training.}
        \label{tab:dmr}
        \resizebox{\linewidth}{!}{
            \begingroup
            \setlength{\tabcolsep}{6pt} 
            \renewcommand{\arraystretch}{1.3} 
            \begin{tabular}{lc}
                \toprule
                \multicolumn{1}{c}{\textbf{Domain Randomization}} & \multicolumn{1}{c}{\textbf{Sampling Distribution}} \\
                \midrule
                Static friction         & $\mathcal{U}(0.3, 1.6)$   \\
                Dynamic friction        & $\mathcal{U}(0.3, 1.2)$   \\
                Restitution             & $\mathcal{U}(0, 0.5)$     \\
                Default joint position  & $\mathcal{U}(-0.01, 0.01)$\\
                Center of mass          & $\mathcal{U}(-0.05, 0.05)$\\
                Random push interval    & $\mathcal{U}(1.0, 3.0)$   \\
                Random push linear velocity & $v_x:\mathcal{U}(-0.5, 0.5)$  \\
                 & $v_y:\mathcal{U}(-0.5, 0.5)$  \\
                 & $v_z:\mathcal{U}(-0.2, 0.2)$  \\
                Random push angular velocity    & $\omega_r:\mathcal{U}(-0.52, 0.52)$    \\
                 & $\omega_p:\mathcal{U}(-0.52, 0.52)$    \\
                 & $\omega_y:\mathcal{U}(-0.78, 0.78)$    \\
                \bottomrule
            \end{tabular}
            \endgroup
        }
    \end{minipage}
\end{figure}

\begin{table}[t!]
\caption{Rewards for tracking policy training.}
\label{tab:rew}
\centering
\resizebox{0.73\linewidth}{!}
{
\begingroup

\setlength{\tabcolsep}{10pt} 
\renewcommand{\arraystretch}{1.3} 
\begin{tabular}{lcr}
        \toprule
        \multicolumn{1}{c}{\textbf{Term}} & \multicolumn{1}{c}{\textbf{Expression}} & \multicolumn{1}{c}{\textbf{Weight}} \\
        \midrule
        \textbf{Task Reward} & & \\
        \hspace{2em} Root position       & $\exp(-\|x_t - x_t^{\rm ref}\| / 0.09)$   & 0.5   \\
        \hspace{2em} Root rotation       & $\exp(-\|r_t - r_t^{\rm ref}\| / 0.16)$   & 0.5   \\
        \hspace{2em} DoF position        & $\exp(-\|q_t - q_t^{\rm ref}\| / 0.16)$   & 1.0   \\
        \hspace{2em} DoF velocity        & $\exp(-\|\dot{q}_t - \dot{q}_t^{\rm ref}\| / 1.00)$    & 1.0   \\
        \hspace{2em} Keypoint position   & $\exp(-\|p_t - p_t^{\rm ref}\| / 0.09)$   & 1.0   \\
        \hspace{2em} Keypoint linear velocity    & $\exp(-\|\dot{p}_t - \dot{p}_t^{\rm ref}\| / 1.00)$  & 1.0 \\
        \textbf{Penalty} & & \\
        \hspace{2em} Action rate         & $-\|a_t - a_{t-1}\|$                      & 0.1   \\
        \hspace{2em} DoF limit           & $-\mathbb{I}(q_t \notin [q_{\rm min}, q_{\rm max}])$  & 10.0 \\ 
        \hspace{2em} Undesired contacts  & $-\sum_{c \notin \{{\rm ankles, wrists}\}}\mathbb{I}(\|\mathbf{F}_c\| > 1.0N)$    & 0.1 \\
        \bottomrule
    \end{tabular}
\endgroup
}
\end{table}

\subsection{Dataset Composition}
\label{app:data}

We construct our language–action dataset by tracking human motion sequences using the trained whole body control policy. We use a subset of the policy training data whose motion sequences are annotated with language descriptions in HumanML3D \citep{Guo_2022_CVPR}, along with their mirrored version, resulting in 12,422 motion clips, each paired with three to four textual descriptions. For each motion sequence, we roll out the trained policy 20 times under domain randomization to obtain diverse robot trajectories corresponding to the same target motion. The domain randomization applied during data collection includes those for policy traning, as listed in \Cref{tab:dmr}, along with action noise sampled form $\mathcal{U}(-0.02, 0.02) $ and added to the target joint position $q_t^{\rm target}$. To ensure dataset quality, we retain only successful trajectories when constructing the final language–action dataset $\mathcal{D}_{\rm robot}$.

\section{Language-Action Model}
\label{app:model}

\subsection{Model Architecture}

Our model builds upon the architecture of SmolVLA \citep{shukor2025smolvla,cadene2024lerobot}, with several modifications tailored for text-based humanoid whole-body control.

We adopt the architecture of SmolVLA as our backbone, but we do not initialize the model with any pretrained VLM or VLA weights. Instead, we train it from scratch on our language–action dataset. In addition, we omit the vision encoder and retain only the language-model component of the architecture for text conditioning and state history encoding. For the 600M model (our default model in experiments), we use \texttt{SmolVLM-Base} with num\_vlm\_layers set to 6. For the 200M and 60M variants, we adopt \texttt{SmolVLM2-500M-Video-Instruct} and \texttt{SmolVLM2-256M-Video-Instruct}, respectively, each configured with num\_vlm\_layers set to 12. We do not adopt layer skipping for the action expert, so the action expert shares the same number of layers as the backbone.

We first encode the language command using the CLIP text encoder \citep{radford2021learning} to obtain semantic token embeddings. Both the text embeddings and the proprioceptive state inputs are then projected into the hidden dimension of the language–state encoder through separate linear layers. To distinguish between modalities, we add learnable type embeddings to the projected language and state tokens.
Since the state history is provided at multiple temporal granularities, as defined in \Cref{equ:state_input}, we additionally incorporate learnable time embeddings for the state tokens, enabling the model to capture their temporal relationships. The action expert employs one linear projector to embed the sampled action noise into its hidden dimension, and another to decode the expert’s hidden states back into the action space.

The state history consists of short-term states $\mathbf{s}_t^{\rm short\_term}$ sampled at 50 Hz and long-term states $\mathbf{s}_t^{\rm long\_term}$ sampled at 4 Hz, defined as:
\begin{align}
    \mathbf{s}_t^{\rm short\_term} &= [s_{t-0.18}, s_{t-0.16}, \ldots, s_{t-0.02}, s_{t}], \\
    \mathbf{s}_t^{\rm long\_term} &= [s_{t-10}, s_{t-9.75}, ..., s_{t-0.5}, s_{t-0.25}].
\end{align}
Such a multi-scale observation preserves high-frequency control feedback while exposing long-horizon temporal structure to the model.

The done prediction head is a two-layer MLP with a hidden dimension same as that of the language–state encoder. It takes as input the the last hidden state of the final state token (\textit{i.e.}, $s_t$) and outputs a scalar indicating the probability that the task will be completed within the next $H$ steps, \textit{i.e.}, the length of the action chunk.

\subsection{Training Details}

We train our language–action model using the AdamW \citep{loshchilov2019decoupled} optimizer with a batch size of 1024 distributed across 4 NVIDIA A800-SXM4-80GB GPUs. The initial learning rate is set to 1e-4, with a cosine decay schedule applied over 100k gradient update steps to 2.5e-6. During training, we randomly mask language inputs with a probability of 0.1 to facilitate classifier-free guidance. The model is trained for 100k gradient update steps, which takes approximately 24 hours.

The training objective consists of two components: a primary flow matching loss and a done prediction loss. The flow matching loss $L_{\rm fm}$ is defined in \Cref{equ:fm}. The done prediction loss $L_{\rm done}$ is a binary cross-entropy loss that encourages the model to estimate whether the task will terminate within the next action chunk. The overall loss is:
\begin{equation}
    L = L_{\rm fm} + \lambda L_{\rm done},
\end{equation}
where $\lambda = 0.01$. To prevent interference with humanoid control learning, gradients from the done prediction head are not propagated back into the language–state encoder. Thus, $L_{\rm done}$ updates only the parameters of the done prediction head.

\subsection{Classifier-Free Guidance}

\begin{wrapfigure}{r}{0.4\textwidth} 

    \centering
    \vspace{-15pt} 

    \includegraphics[width=\linewidth]{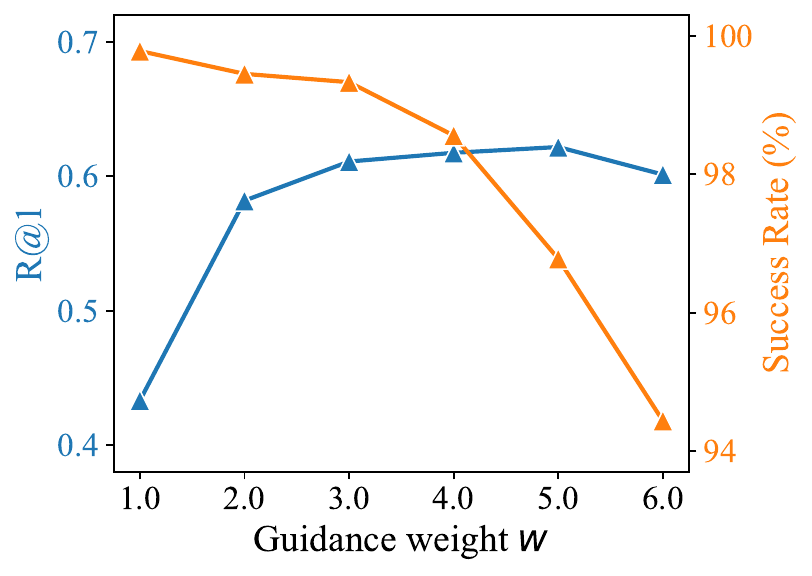}
    
    \caption{Effect of classifier-free guidance weight $w$.}
    \label{fig:cfg}
    
    \vspace{-4em}
\end{wrapfigure}
During inference, we apply classifier-free guidance to enhance the alignment between generated actions and language commands. Specifically, we perform two forward passes through the model at each denoising step: one with the actual language input and another with the language input masked out. The final predicted action noise is computed as a weighted combination of the two outputs:
\begin{align}
    v_{\rm cond} &= v_{\theta}(\mathbf{A}_t^\tau,\tau,[l,\mathbf{s}_t^{\rm hist}]) \\
    v_{\rm uncond} &= v_{\theta}(\mathbf{A}_t^\tau,\tau,[\varnothing,\mathbf{s}_t^{\rm hist}]) \\
    v_{\rm cfg} &= v_{\rm uncond} + w(v_{\rm cond} - v_{\rm uncond}) \\
    \mathbf{A}_t^{\tau - \Delta t} &= \mathbf{A}_t^{\tau} - v_{\rm cfg} \cdot \Delta t.
\end{align}
Here, $w > 1$ is the guidance weight that amplifies the influence of the language command on the generated actions. We conduct an experiment on the guidance weight as shown in \Cref{fig:cfg}. The R@1 improves as we increase $w$ from 1.0 to 5.0, indicating better alignment between the generated actions and language commands. However, the success rate decreases with larger $w$, suggesting that overly aggressive guidance may lead to less stable control. This indicates a trade-off between physical fidelity and language alignment when tuning the guidance weight for text-based whole body control. We choose $w=2.0$ as a balanced setting for our main experiments.

\subsection{Residual Post-Training}
\label{app:posttrain}

\begin{table}[t]
\caption{Rewards for residual post-training.}
\label{tab:res_rew}
\centering

\resizebox{0.7\linewidth}{!}
{
\begingroup

\setlength{\tabcolsep}{10pt} 
\renewcommand{\arraystretch}{1.3} 
\begin{tabular}{lcr}
        \toprule
        \multicolumn{1}{c}{\textbf{Term}} & \multicolumn{1}{c}{\textbf{Expression}} & \multicolumn{1}{c}{\textbf{Weight}} \\
        \midrule
        \textbf{Task Reward} & & \\
        \hspace{2em} DoF position        & $\exp(-\|q_t - \hat{q}_t\| / 0.16)$  & 2.0   \\
        \hspace{2em} Active termination  & 1.0 if LA model predicts done   & 100.0 \\
        \textbf{Penalty} & & \\
        \hspace{2em} Residual action norm& $\exp(-\|\Delta a_t\| / 0.09)$                        & 2.0   \\
        \hspace{2em} Action rate         & $-\|a_t^{\rm final} - a_{t-1}^{\rm final}\|$          & 0.2   \\
        \hspace{2em} DoF limit           & $-\mathbb{I}(q_t \notin [q_{\rm min}, q_{\rm max}])$  & 10.0  \\
        \hspace{2em} Undesired contacts  & $-\sum_{c \notin \{{\rm ankles, wrists}\}}\mathbb{I}(\|\mathbf{F}_c\| > 1.0N)$    & 0.1 \\
        \hspace{2em} Termination         & -1.0 if fall down                          & 100.0  \\
        \bottomrule
    \end{tabular}
\endgroup
}
\end{table}

To train $\pi_\Delta$, we freeze the parameters of the language–action model and optimize the residual head with PPO \citep{schulman2017proximal} in IsaacLab \citep{mittal2023orbit}. We apply domain randomization over physical parameters and external perturbations to improve the robustness of our method, same as those used during the tracking policy training, as listed in \Cref{tab:dmr}. The residual action head $\pi_\Delta$ is implemented as an MLP with three hidden layers of sizes [512, 256, 128], and is trained with the same PPO hyperparameters listed in \Cref{tab:ppo}, except that the number of parallel environments is reduced to 1024 due to GPU memory limits. Before being added to the original action predicted by the language–action model, the residual action $\Delta a_t$ is scaled by a factor of 0.02.

The reward functions as summarized in \Cref{tab:res_rew}. The DoF position tracking term encourages the residual action to drive the future joint positions toward the original predictions $\hat{q}$ under domain randomization. The active termination means that the language-action model predicts the task is done, which encourages the residual policy to complete the task as intended. The residual action norm term penalizes large residual corrections to prevent the policy from deviating excessively from the original action.

\begin{wrapfigure}{r}{0.5\textwidth} 
    \vspace{-4mm}
    \captionof{table}{Ablation study results for residual post-training.}
    \label{tab:ablation_res}
    \centering
    \resizebox{1.0\linewidth}{!}
    {
    \begingroup
    
    \setlength{\tabcolsep}{7pt} 
    \renewcommand{\arraystretch}{1.3} 
    \begin{tabular}{lccc}
    \toprule
    \textbf{Method} & R@1 $\uparrow$ & MMD $\downarrow$ & Success $\uparrow$ \\
    \midrule
    \textbf{Base + $\pi_\Delta$} & \textbf{0.392} & \textbf{6.100} & 99.11 \\
    \textbf{Base + $\pi_\Delta$ w/o $r^{\rm DoF}$} & 0.086 & 25.358 & \textbf{99.89} \\
    \textbf{Base + $\pi_\Delta$ w/o $r^{\rm Done}$} & 0.255 & 9.779 & 99.78 \\
    \bottomrule
    \end{tabular}
    \endgroup
    }
\end{wrapfigure}

To evaluate the effectiveness of the two task reward terms, \textit{i.e.}, DoF (joint) position tracking $r^{\rm DoF}$ and active termination $r^{\rm Done}$, we conduct an ablation study during residual post-training. As shown in \Cref{tab:ablation_res}, removing active termination results in a significant reward hacking, where the residual action head learns to output corrections that make the humanoid robot control overly conservative, leading to a high success rate but very poor language following. In practice, we observe that the robot tends to maintain a stable standing pose without performing any movements, simply to avoid the large penalty from falling. Similarly, removing the DoF position tracking term also leads to conservative behavior, albeit to a lesser extent than the removal of active termination.

\section{Experiment Details}
\label{app:experiments}

\subsection{Baselines}
\label{app:baseline}

\noindent \textbf{MDM/T2M-GPT + Retarget.} We use the official checkpoints of MDM \citep{tevet2023human} and T2M-GPT \citep{zhang2023generating} to generate human motion sequences from our test language commands. The generated motions are then retargeted to humanoid robot kinematics using the same retargeting pipeline as in our tracking policy training. The resulting robot motions are executed using our tracking policy.

\vspace{1mm}
\noindent \textbf{UH-1.} We reproduce UH-1 \citep{mao2025universal} based on the T2M-GPT codebase, but replacing human motion with humanoid robot motion obtained by the same retargeting pipeline as in our tracking policy training. The generated robot motion is subsequently executed via our tracking policy.

\vspace{1mm}
\noindent \textbf{LangWBC.} We implement LangWBC \citep{shao2025langwbc} within our whole body control policy training framework and distill the CVAE-based policy using our tracking policy as the teacher model.

\vspace{1mm}
We use the same tracking policy for motion execution (MDM/T2M-GPT + Retarget and UH-1), policy distillation (LangWBC), and BC data collection (ours) to ensure a fair comparison, where variations arising from the underlying whole-body controller are effectively removed.

\subsection{Evaluation Metrics}
\label{app:metrics}

We train a TMR \citep{petrovich23tmr} on our language-action dataset to learn the alignment between language commands and real robot trajectories. Specifically, each trajectory is downsampled to 10Hz, and the input for each frame consists of (i) root linear velocity and root angular velocity, (ii) joint positions and joint velocities, and (iii) the local translations and 6D rotations \citep{hempel20226d} of 29 rigid bodies. 

Based on the trained TMR and collected evaluation trajectories $\mathcal{D}_{\rm robot}^{\rm eval} = \{(\boldsymbol{\tau}_i, l_i)\}_{i=1}^N$, we first compute a similarity matrix $S \in \mathbb{R}^{N \times N}$, where each element is defined as:
\begin{equation}
    S_{i,j} = \frac{f_{\rm TMR}^{\rm t}(l_i)^\top f_{\rm TMR}^{\rm m}(\boldsymbol{\tau}_j)}{\|f_{\rm TMR}^{\rm t}(l_i)\| \|f_{\rm TMR}^{\rm m}(\boldsymbol{\tau}_j)\|}.
\end{equation}
Here, $f_{\rm TMR}^{\rm t}(\cdot)$ denotes the text encoder of TMR, and $f_{\rm TMR}^{\rm m}(\cdot)$ represents the motion encoder of TMR. Using the similarity matrix $S$, we compute the following evaluation metrics:

\vspace{1mm}
\noindent \textbf{Multi-Modal Distance.} This metric measures the average distance between the encoded language command and the encoded robot trajectory in the TMR embedding space:
\begin{equation}
    \text{MM-Dist} = \frac{1}{N} \sum_{i=1}^N \left( 1 - S_{i,i} \right).
\end{equation}

\vspace{1mm}
\noindent \textbf{R-Precision at K.} This metric evaluates the retrieval accuracy of matching language commands to robot trajectories. Following \citep{Guo_2022_CVPR}, for each language command $l_i$, we rank a batch of 32 robot trajectories, including the paired one, based on their similarity scores $S_{i,j}$. R-Precision at K (R@K) is defined as the proportion of language commands for which the correct trajectory $\boldsymbol{\tau}_i$ is among the top K retrieved trajectories.

\vspace{1mm}
\noindent \textbf{Diversity.} This metric measures the variability of generated robot trajectories in the TMR embedding space. Following \citep{Guo_2022_CVPR}, it is computed as the average pairwise distance between two randomly sampled trajectory subset from the evaluation set. Formally, given two subsets $\mathcal{A}, \mathcal{B} \subset \mathcal{D}_{\rm robot}^{\rm eval}$, each containing $N_s = 300$ trajectories, the diversity is defined as:
\begin{equation}
    \text{Diversity} = \frac{1}{N_s} \sum_{i=1}^{N_s} (1 - S_{a_i, b_i}),
\end{equation}
where $a_i$ and $b_i$ are the indices of the $i$-th trajectory in subsets $\mathcal{A}$ and $\mathcal{B}$, respectively. We report the average diversity over 10 random pairs of subsets.

\vspace{2mm}
\noindent \textbf{Maximum Mean Discrepancy.} This metric measures the distributional difference in the TMR embedding space between the generated robot trajectories $\{\boldsymbol{\tau}_i\}_{i=1}^N$ and the ground-truth robot trajectories $\{\boldsymbol{\tau}_i^{\rm GT}\}_{i=1}^M$ collected by our tracking policy. Denote the TMR encoded trajectories as $e_i = f_{\rm TMR}^{\rm m}(\boldsymbol{\tau}_i)$ and $e_i^{\rm GT} = f_{\rm TMR}^{\rm m}(\boldsymbol{\tau}_i^{\rm GT})$. Following \citep{jayasumana2024rethinking}, we compute MMD as:
\begin{align}
    \text{MMD} = \frac{1}{N(N-1)} \sum_{i=1}^N \sum_{j \neq i}^N k(e_i, e_j) + \frac{1}{M(M-1)} \sum_{i=1}^M \sum_{j \neq i}^M k(e_i^{\rm GT}, e_j^{\rm GT}) - \frac{2}{NM} \sum_{i=1}^N \sum_{j=1}^M k(e_i, e_j^{\rm GT}),
\end{align}
where $k(x, y) = \exp(-\|x-y\|^2 / 2\sigma^2)$ is the Gaussian RBF kernel with bandwidth $\sigma = 10$. We scale up the MMD value by a factor of 100 for better readability.

\subsection{Waypoint navigation}
\label{app:navi}

To construct a navigation-oriented subset of our language–action dataset, we design a set of 200 command templates that specify target positions in a 2D plane. The templates are structured to describe a person walking to a specific coordinate, with variations in phrasing to enhance linguistic diversity. These templates are generated using ChatGPT. Examples of these templates are provided below:
\vspace{2mm}
\begin{mdframed}
\textbf{Template examples:} \\
    ``A person walks to position (\textit{x} unit, \textit{y} unit)." \\
    ``Someone moves toward x = \textit{x} unit, y = \textit{y} unit." \\
    ``A man goes to location (\textit{x} unit, \textit{y} unit)." \\
    ``A woman walks toward (\textit{x} unit, \textit{y} unit)." \\
    ``A human steps to (\textit{x} unit, \textit{y} unit)." \\
    ``An individual proceeds to coordinate (\textit{x} unit, \textit{y} unit)." \\
    ``A pedestrian walks toward x = \textit{x} unit, y = \textit{y} unit." \\
    ``A man strides to (\textit{x} unit, \textit{y} unit)." \\
    ``A woman heads toward (\textit{x} unit, \textit{y} unit)."
\end{mdframed}
\vspace{2mm}
Based on the displacements from the starting frame to the ending frame of approximately 2,000 walking-type motions in the AMASS dataset \citep{AMASS:ICCV:2019}, we generate movement commands by filling in the target coordinates in the templates, using 20 cm as the discretization unit. For instance, if a motion sequence involves a displacement of (1.2 m, 0.8 m), we would create a command like ``\textit{A person walks to position (6 unit, 4 unit)}". We randomly apply three templates to each trajectory and finetune our language-action model on this dataset. 

To quantitatively evaluate waypoint tracking accuracy, we randomly sample target positions within the range of 4m to 6m along the x-axis and -2m to 2m along the y-axis, and conduct a two-round navigation task. A navigation round is considered successful if the robot reaches within 0.5m of the target position and will skip next round. For the second iteration, we transform and discretize the target relative to the robot's frame at the end of the first round. For each navigation round, we sample one of the templates and fill in the discretized target coordinates to form the language input.

\subsection{Real World Deployment}
\label{app:real}

\begin{figure}
    \centering
    \includegraphics[width=0.85\linewidth]{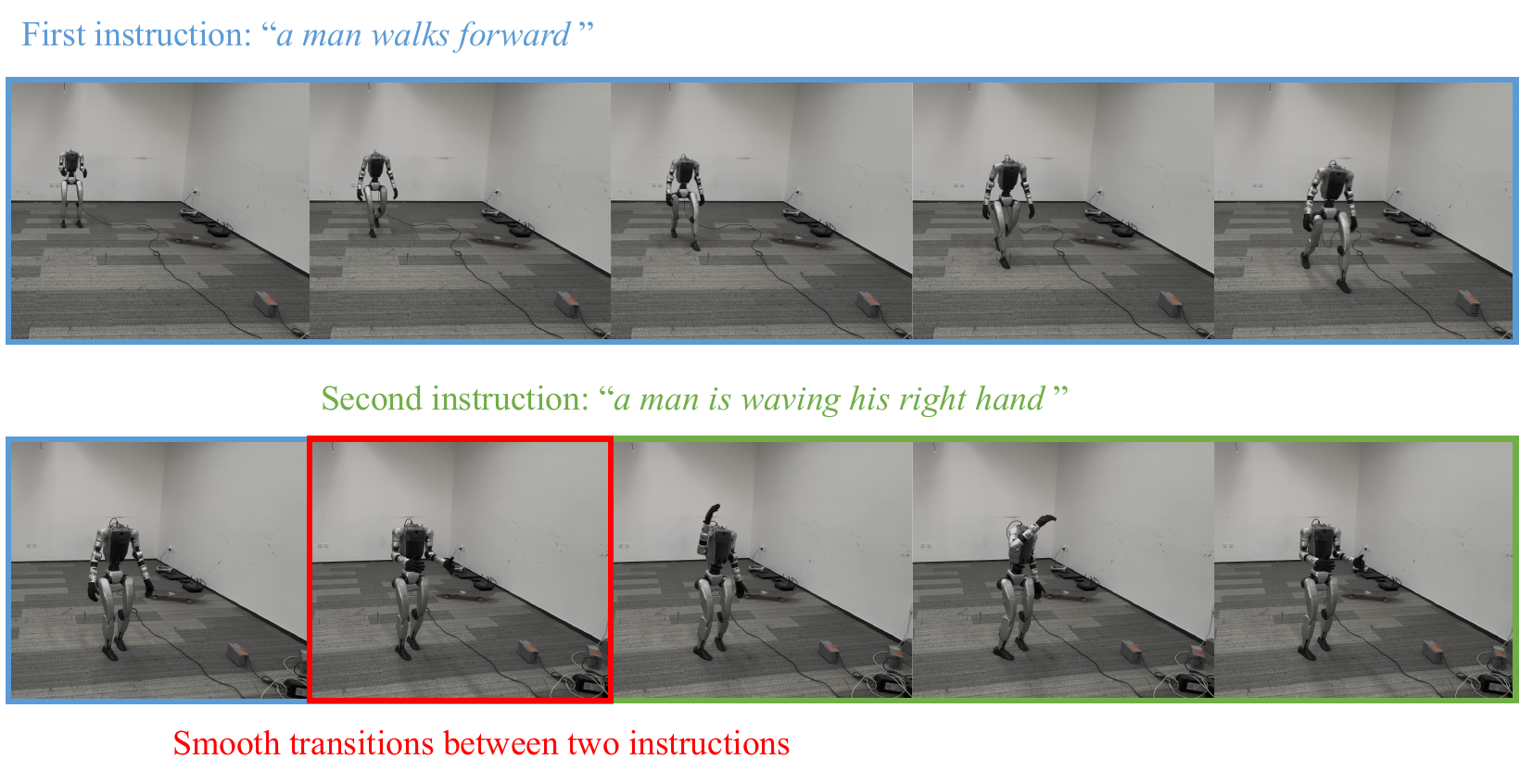}
    \caption{Autonomous sequential execution of two instructions: ``\textit{a man walks forward}" and ``\textit{a man is waving his right hand}". Our system produces smooth, continuous transitions between two instructions.}
    \label{fig:real_demo}
\end{figure}

For real-world deployment, we use the same Unitree G1 humanoid robot as in simulation. The language–action model runs on an NVIDIA GeForce RTX 4090 GPU, which communicates with the robot through a wired connection. Proprioceptive states are streamed from the robot to the GPU at 50 Hz, where the model performs action generation via flow matching. The predicted actions are then returned to the onboard PD controller, operating at 200 Hz.

Unlike in simulation, where physics can be paused while waiting for the next inference step, the real robot continues to execute the last action during model computation. To accommodate this, we adopt an asynchronous inference pipeline \citep{shukor2025smolvla}, which decouples action generation from the control loop. The policy produces an action chunk of length $K_1$ every $K_2$ steps, with $K_1 > K_2$. Because the action expert is a causal transformer, it can generate shorter action chunks during evaluation than the full training horizon $H$ used for training. While a new chunk is being generated, the robot continues executing the unconsumed actions from the previous chunk, ensuring uninterrupted control.

After a new action chunk becomes available, the system discards the first $K_3$ actions, corresponding to the steps elapsed during model inference, and replaces the previous action chunk with the remaining actions. Here, $K_3 = \lceil {T_{\rm infer}}/{0.02} \rceil$, with $T_{\mathrm{infer}}$ denoting the model’s inference latency and 0.02 s corresponding to the 50 Hz control cycle. In our experiments, we use $K_1 = 6$ and $K_2 = 4$, matching the rollout step used in our simulation experiments.

\begin{wraptable}{r}{0.5\columnwidth}
\vspace{-4mm}
\caption{Inference time of our model on an NVIDIA GeForce RTX 4090 GPU.}
\label{tab:time}
\centering
\resizebox{\linewidth}{!}
{
\begingroup

\setlength{\tabcolsep}{8pt} 
\renewcommand{\arraystretch}{1.3} 
\begin{tabular}{lc}
\toprule
\textbf{Model Part} & \textbf{Inference Time} \\
\midrule
LA model with 10 flow match pass & 26ms \\
LA model with 5 flow match pass  & 16ms \\
Communication overhead           & 2ms  \\
\bottomrule
\end{tabular}
\endgroup
}
\end{wraptable}

As shown in \Cref{tab:time}, a single inference of the language–action model takes 26 ms using 10 flow matching steps and 16 ms with 5 steps. Since the latter provides comparable performance in real-world tests, we adopt the 5-step flow matching for most hardware experiments to reduce latency. The communication overhead between the robot and the GPU is approximately 2 ms. Overall, the asynchronous inference pipeline effectively compensates for model latency and enables smooth and robust whole-body control on real humanoid hardware. Besides \Cref{fig:demo}, we provides \href{https://youtu.be/U5B5Pgw1N3A}{real world deployment videos} to demonstrate that our model and inference pipeline can reliably execute diverse language instructions.

\vspace{1mm}
\noindent \textbf{Autonomous Sequential Execution.} Building on this capability, our system also supports autonomous sequential execution of multiple language commands. During dataset construction, each motion sequence is augmented with smooth transitions to and from a default standing pose. As a result, the language–action model learns a consistent behavioral pattern: it initiates each command from the default pose and returns to the same pose after completion\footnote{To ensure fair comparison, we apply the same preprocessing to all baseline methods in previous experiments.}. This default pose anchoring enables the robot to execute multiple commands consecutively without manual reset: after returning to the default standing pose, the model’s done prediction is used to autonomously terminate the current command and proceed to the next one, which again begins from the same default pose. Consequently, the system produces smooth, continuous transitions between behaviors and supports long-horizon control, as shown in \Cref{fig:real_demo}.

\newpage
\section{Credit Authorship}

\begin{multicols}{2}

\noindent\textbf{Initial Exploration}
\begin{itemize}
    \item \textsc{Haobin Jiang}
    \item \textsc{Ziluo Ding}
\end{itemize}

\noindent\textbf{Whole Body Controller Training}
\begin{itemize}
    \item \textsc{Ziluo Ding}
    \item \textsc{Yuxuan Wang}
\end{itemize}

\noindent\textbf{Data Generation \& Training}
\begin{itemize}
    \item \textsc{Yuxuan Wang}
    \item \textsc{Haobin Jiang}
\end{itemize}

\columnbreak

\noindent\textbf{Post Training}
\begin{itemize}
    \item \textsc{Shiqing Yao}
    \item \textsc{Haobin Jiang}
\end{itemize}

\noindent\textbf{Real-World Deployment}
\begin{itemize}
    \item \textsc{Yuxuan Wang}
    \item \textsc{Haobin Jiang}
\end{itemize}

\noindent\textbf{Project Lead}
\begin{itemize}
    \item \textsc{Ziluo Ding}
    \item \textsc{Zongqing Lu}
\end{itemize}
\end{multicols}

\clearpage

\end{document}